\documentclass{midl} % Anonymized submission

\usepackage{mwe} % to get dummy images
\usepackage{bm} % for bold maths symbols
\usepackage{bbm}
\usepackage{xargs}  
\jmlrvolume{-- Under Review}
\jmlryear{2024}
\jmlrworkshop{Full Paper -- MIDL 2024 submission}
\editors{Under Review for MIDL 2024}

\title[The Role of Bilateral Symmetry in Brain Inpainting]{Investigating the Role of Bilateral Symmetry for Inpainting Brain MRI}

\midlauthor{
\Name{Sergey Kuznetsov} \Email{sk768@sussex.ac.uk}\\
\Name{Sanduni Pinnawala} \Email{m.pinnawala@sussex.ac.uk}\\
\Name{Peter A. Wijeratne} \Email{p.wijeratne@sussex.ac.uk}\\
\Name{Ivor J. A. Simpson}\Email{i.simpson@sussex.ac.uk}\\
\addr Department of Informatics, University of Sussex, UK
}

\newcommand{\structsSet}{\mathcal{S}}
\newcommand{\sidesLeft}{\mathrm{left}}
\newcommand{\sidesRight}{\mathrm{right}}
\newcommand{\sidesIdx}{k}
\newcommand{\mask}{\bm{M}}
\newcommand{\image}{\bm{I}}
\newcommand{\logdetJ}{\log|\bm{J}|}
\newcommand{\imageDist}{\bm{d}}
\newcommand{\identity}{\mathbb{I}}
\newcommandx{\Todo}[2][1=]{\textcolor{red}{\textbf{TODO:} #2}}

\begin{document}

\maketitle

\begin{abstract}
Inpainting has recently emerged as a valuable and interesting technology to employ in the analysis of medical imaging data, in particular brain MRI. A wide variety of methodologies for inpainting MRI have been proposed and demonstrated on tasks including anomaly detection.
In this work we investigate the statistical relationship between inpainted brain structures and the amount of subject-specific conditioning information, i.e. the other areas of the image that are masked. 
In particular, we analyse the distribution of inpainting results when masking additional regions of the image, specifically the contra-lateral structure.
This allows us to elucidate where in the brain the model is drawing information from, and in particular, what is the importance of hemispherical symmetry?
Our experiments interrogate a diffusion inpainting model through analysing the inpainting of subcortical brain structures based on intensity and estimated area change. 
We demonstrate that some structures show a strong influence of symmetry in the conditioning of the inpainting process. 
\end{abstract}

\begin{keywords}
Inpainting, Brain MRI, Structural symmetry, Diffusion models
\end{keywords}

\section{Introduction}
Inpainting, the process of inserting new image information conditioned on a masked context window, has recently become an interesting methodology in medical image analysis with applications in anomaly detection \cite{graham2023unsupervised, cercea2023reversing} and local distortion removal \cite{armanious2019adversarial, armanious2020ipa}. These are commonly formulated using a generative model, which describes the joint distribution of voxel intensities and enables this to be conditioned on specified spatial regions.

In this work, we contribute a novel framework for interrogating the importance of regions of conditioning pixels when inpainting brain MRI, and in particular investigate the role of lateral symmetry to help explain why a particular result may have been derived.

We hypothesise that if lateral symmetry is important, we would expect bilateral masking to lead to higher variability whereas the baseline approach will provide an inpainting that increases the symmetry of appearance leading to more consistent differences in volume and appearance from the original image.

\section{Inpainting in Brain MRI}

The objective of image inpainting is to reconstruct the missing regions of an image by leveraging the information from the surrounding known pixels.

Inpainting can be described as learning a generative model for the image data $p(\image)$ which can be conditioned on arbitrary subsets of pixels $p(\image | \image \circ \mask)$, where $\mask$ describes a binary pixel mask that is 1 for observed pixels and 0 for unobserved. 

In this work we investigate the effect of increasing the number of unknown pixels on the resulting model as a means of introspecting what the generative model has learned about the population of brain imaging data. 

\subsection{Related Work}
Early deep learning-based approaches to inpainting employed encoder-decoder architectures, incorporating both adversarial loss from Generative Adversarial Networks (GANs) \cite{goodfellow2014generative} and reconstruction loss, typically measured by $L2$ distance \cite{pathak2016context, iizuka2017globally}. These approaches have primarily focused on filling rectangular regions in the center of an image, and subsequent studies were aimed on inpainting images with irregularly shaped missing regions using alternative types of convolutional filters \cite{liu2018image, yu2019free}. However, due to the intrinsic characteristics of GANs, the outputs they produce often lack both diversity and semantic coherence \cite{peng2021generating} and the models can sometimes be unstable to train \cite{salimans2016improved, kodali2017convergence}.

Alternative methodologies utilize Variational Autoencoders (VAEs) \cite{zheng2019pluralistic, zhao2020uctgan}, which define a Gaussian prior to model the distribution of continuous latent variables. By drawing samples from this distribution, VAEs can generate multiple distinct variations of the inpainted regions. A significant limitation of VAEs is their tendency to generate outputs that may appear distorted or blurry due to inherent trade-offs in the model's objective function.

Additionally, VAEs can encounter a problem known as ``posterior collapse'' \cite{van2017neural}, where the model starts ignoring hidden hidden variations in the data and relies too heavily on the decoder's ability to generate data based on the prior distribution. As a result, latent variables which should encode meaningful differences, do not contribute significantly to the generation process. This issue restricts the VAEs ability to produce varied outputs, as the model fails to leverage the full potential of its latent space. \cite{peng2021generating} addresses this problem by introducing a Hierarchical Vector Quantized Variational Autoencoder (VQ-VAE) that enables autoregressive distribution over discrete latent variables and demonstrates significant improvements over conventional VAEs on the inpainting task.

Diffusion models \cite{ho2020denoising} are a class of likelihood-based generative models that serve as alternatives to GAN and VAE-based methods, recently proving to be exceptionally effective in image generation tasks \cite{ramesh2022hierarchical, saharia2022photorealistic}. These models operate on the principle of gradually refining the image based on previous estimates, which leads to more consistent training and greater coverage \cite{li2022sdm}. By using a diffusion model in this study, we aim to better understand the variability between different runs of the inpainting algorithm.

\subsection{Use of Diffusion Models}\label{sec:diffusions}

Diffusion models are designed to learn the underlying distribution of a dataset by reversing a gradual denoising process applied to the training images, thereby restoring the original data distribution from a series of increasingly noisy representations. At each timestep $t$, the data point $\mathbf{x}_{t}$ represents a mixture of the original signal $\mathbf{x}_{0}$ and a noise element $\epsilon$. The proportion of the original signal to noise at each timestep is modulated by a value of $t$. A single step of the forward diffusion process can be described as follows:

\begin{align}
q(\mathbf{x}_{t}|\mathbf{x}_{t-1}) = \mathcal{N}(\mathbf{x}_{t}; \sqrt{1 - \beta}\mathbf{x}_{t-1}, \beta_{t}\identity)
\end{align}

Where $\beta$ is a sequence of values $(\beta_{1}, \beta_{2}, \dots, \beta_{T})$ that are used to scale the sample and gradually increase the amount of noise at each timestep of the diffusion process. We can define the forward process for a sample $\mathbf{x}_0$ at a time step $t$ in a closed form by introducing $\alpha_{t} = 1 - \beta_{t}$ and $\bar{\alpha}_{t} = \Pi_{s=1}^t \alpha_{s}$

The diffusion model is trained to reverse this process and predict a slightly less noisy version of the sample $\mathbf{x}_{t-1}$ from $\mathbf{x}_{t}$:

\begin{align}
p_{\theta}(\mathbf{x}_{t-1}|\mathbf{x}_{t}) = \mathcal{N}\bigl(\mathbf{x}_{t-1}|\mu_{\theta}(\mathbf{x}_{t}, t), \Sigma_{\theta}(\mathbf{x}_{t}, t)\bigr)
\end{align}

Where $\Sigma_{\theta}$ typically remains fixed and the training focuses on predicting the mean $\mu_{\theta}$ based on the noisy sample $\mathbf{x}_{t}$ and specific timestep $t$.

The objective function can be derived by specifying the variational lower bound over the underlying data distribution:

\begin{align}
    \mathbb{E}[- \log p_{\theta}(\mathbf{x}_{0})] &\leq \mathbb{E}_{q}\Bigl[ - \log p_{\theta}(\mathbf{x}_{0}) + D_{KL} \bigl(q(\mathbf{x}_{1:T}|\mathbf{x}_{0})|| p_{\theta}(\mathbf{x}_{1:T}|\mathbf{x}_{0})\bigr)\Bigr] \\
    &\leq \mathbb{E}_{q}\Bigl[\sum_{t=2}^T \underbrace{D_{KL}\bigl(q(\mathbf{x}_{t-1}|\mathbf{x}_{t}, \mathbf{x}_{0})||p_{\theta}(\mathbf{x}_{t-1}|\mathbf{x}_{t})\bigr)}_{\| \epsilon_{t} - \epsilon_{\theta}(\mathbf{x}_{t}, t) \|^2} - \log \big(p_{\theta}(\mathbf{x}_{0}|\mathbf{x}_{1})\bigr)\Bigr]
\end{align}

According to \cite{ho2020denoising}, the model can be parameterized to predict the noise component which is then subtracted from the noisy sample:

\begin{align}
\mu_{\theta}(\mathbf{x}_{t}, t) = \frac{1}{\sqrt{\alpha}}\Bigl(\mathbf{x}_{t} - \frac{\beta_{t}}{\sqrt{1 - \bar{\alpha_{t}}}} \epsilon_{\theta}(\mathbf{x}_{t}, t) \Bigr)
\end{align}

The objective can be further simplified to:
\begin{align}
    \mathcal{L}_{t}^{simple} = \mathbb{E}_{t \sim [1, T], \mathbf{x}_{0}, \epsilon_{t}} \Bigl[ \| \epsilon_{t} - \epsilon_{\theta} (\sqrt{\bar{\alpha}_{t}}\mathbf{x}_{0} + \sqrt{1 - \bar{\alpha}_{t}}\epsilon_{t}) \|^2 \Bigr]
\end{align}

\section{Methods}
\subsection{Datasets}
In our study, we utilized two primary datasets. The first, \textbf{IXI}\footnote{\url{https://brain-development.org/ixi-dataset/}} dataset, encompasses approximately 600 MRI scans sourced from healthy individuals. These MRI scans were subsequently sliced and segmented into nearly 28000 2D slices, where 90\% of these slices were used for model training and the remaining 10\% were allocated for validation. The other one is the \textbf{HCP}\footnote{\url{https://www.humanconnectome.org/study/hcp-young-adult}} dataset, which includes high-resolution MRI scans of young, healthy adults aged 22 to 35 years. For our experiments, we chose five subjects from the dataset and performed the inpainting process five times on each subject to evaluate the consistency of the inpainting algorithm across different trials.

At the preprocessing stage, our primary goal was to align the data distributions between the two datasets as closely as possible. All MRI volumes were registered to the FreeSurfer's FsAverage, ensuring spatial alignment, followed by white matter normalization. The extracted 2D slices were then cropped and resized to align with the input size requirements of a pretrained UNet model. The corresponding masks were cropped similarly and inverted to meet the specific formatting needs of our chosen inpainting algorithm. 

To further harmonize the distributions between the two datasets, histogram matching was applied. Finally, the inpainted slices are concatenated together to reconstruct the volumetric structure of the original brain scan.

\subsection{Choice of Inpainting Models}
The inpainting approach we employ utilizes the RePaint algorithm \cite{lugmayr2022repaint} and involves a binary mask $\mask$ to differentiate between known $(\mask = 1)$ and unknown $(\mask = 0)$ regions in an image $\image$. The algorithm is used at a sampling phase with a pre-trained model and does not require any additional training on its own. During the inpainting, we first sample the masked region $\image_{t}^{\text{known}}$ by applying noise to the original image $\image_{0}$ and then generate unmasked pixels $\image_{t}^{\text{unknown}}$ from the model, which are subsequently combined with the known region (See \figureref{fig:repaint}). A single step of the inpainting process can be delineated as follows:

\begin{align}
&\image_{t-1}^{\text{known}} = \mathcal{N}(\sqrt{\bar{\alpha}_{t-1}}\image_{0}, \bigl(1-\bar{\alpha_{t}})\identity\bigr) \\
&\image_{t-1}^{\text{unknown}} = \mathcal{N}(\mu_{\theta}\bigl(\image_{t}, t), \Sigma_{\theta}(\image_{t}, t)\bigr) \\
&\image_{t-1} = \mask \odot \image_{t-1}^{\text{known}} + (1 - \mask) \odot \image_{t-1}^{\text{unknown}}
\end{align}

Within the inpainting process, each successive step focuses exclusively on the known region for sampling the masked pixel values, which can result in a disharmony between the sampled and the generated image regions. To address this, it is suggested to apply a forward diffusion process from $\image_{t-1}$ back to $\image_{t}$ to preserve some of the information contained in $\image_{t-1}^{\text{unknown}}$ enhancing harmony between the known and generated parts of the image. 

\begin{figure}[htbp]
 % Caption and label go in the first argument and the figure contents
 % go in the second argument
\floatconts
  {fig:repaint}
  {\caption{The schematic of the RePaint algorithm applied to a brain MRI.}}
  {\includegraphics[width=0.9\linewidth]{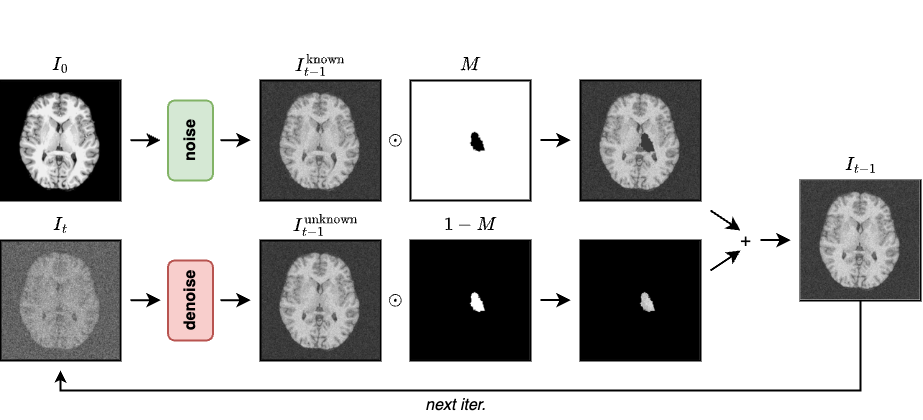}}
\end{figure}

This version of the algorithm introduces two new parameters to control the inpainting process. Namely, the \textit{jump length }$\bm{j}$, which controls the number of forward steps we do to diffuse the image before going back in time again, and the \textit{number of resamplings} $\bm{r}$, which determines the number of times we repeat this process at each chosen timestep.

\subsection{Model Training}

In our research, we have adopted the Denoising Diffusion Probabilistic Model (DDPM), described in \cite{dhariwal2021diffusion}. The foundational model was trained on $256 \times 256$ images from ImageNet dataset \cite{deng2009imagenet}. We perform fine-tuning on the IXI dataset over $10$ epochs starting with a learning rate of $0.001$ and using cosine annealing \cite{loshchilov2016sgdr}. Due to high computational cost of training a diffusion model, we adopt a batch size of 2 with gradient accumulation over 8 batches.

\subsection{Analysis Methodology}
To investigating the influence of lateral symmetry or local information in inpainting, we propose the following procedure:

For each image, $\image$, we have an associated set of binary segmentation masks $\mask^{\sidesIdx}_{s} \in  \mathbb{Z}_2^{h\times w}$ , where $\sidesIdx \in \{\sidesLeft, \sidesRight\}$ and $s \in \mathcal{S}$ where $\structsSet$ is a set of subcortical brain structures, $\mathcal{S} \in \{$amygdala, caudate, hippocampus, lateral\_ventricle, palidum, putamen, thalamus$\}$.

\begin{algorithm2e}
\caption{The experimental procedure for a given image and set of segmentation masks}
\label{alg:net}
 % older versions of algorithm2e have \dontprintsemicolon instead
 % of the following:
 %\DontPrintSemicolon
 % older versions of algorithm2e have \linesnumbered instead of the
 % following:
 %\LinesNumbered
\KwIn{$\image, \mask$}
\KwOut{$\logdetJ$, a tensor of log Jacobian determinant values}
\KwOut{$\bm{d}$, a tensor of image similarity measures}
\For(\tcp*[f]{Iterate over subcortical structures}){ $s \in \structsSet$} 
{
\For{$\sidesIdx \in \{ \sidesLeft, \sidesRight\}$}
{
\lIf{{Baseline Experiment}}{$\mask^* \leftarrow \mask^k_s$} \lElseIf{Bilateral masking}{$\mask^* \leftarrow \mask^k_s \cap \mask^{!k}_s$}
%\ElseIf{Dilated masking}{$\mask^* \leftarrow D(\mask^k_s)$;}

\For{$i \leftarrow 1$ \KwTo $n_s$}{
$\hat{\image}_i \leftarrow \mathrm{inpaint}(\image \circ \mask^*) \circ (\mathbbm{1}-\mask^k_s) + \image \circ \mask^k_s $

\tcp{Use non-linear registration to calculate pixelwise volume change}
$|\bm{J}|^k_{s,i} \ \leftarrow \mathrm{CalculateDetJ}(\mathrm{Registration}(\image, \hat{\image}_i)) \circ (\mathbbm{1}-\mask^k_s)$ 

\tcp{Calculate the image similarity with the inpainted image}
$\mathbf{d}^k_{s,i} \ \leftarrow \mathrm{SIM}(\image, \hat{\image}_i \circ (\mathbbm{1}-\mask^k_s))$
}
}
}
\label{alg1}
\end{algorithm2e}

The algorithm for performing the different inpainting experiments is given in \ref{alg1}. We measure the accuracy of the inpainting using two measures:
\begin{itemize}
    \item $\logdetJ$ which is the log determinant of the Jacobian obtained by non-linear registration of the inpainted image to the original image using Niftyreg  \cite{modat2010fast}. This describes the log-transformed proportional change in area/volume of the inpainted region with respect to the original image. Values of 0 indicate no volume changes, $\log 2$ is a doubling and $\log 0.5$ is halving.
    \item $\imageDist$ which measures the similarity between the inpainted and reference images using the mean squared error within the masked region.
\end{itemize}

Using these two measures, we can estimate the effect of masking additional regions on the consistency of inpainting with the original images.

\begin{figure}
    \centering
    \includegraphics[width=\textwidth]{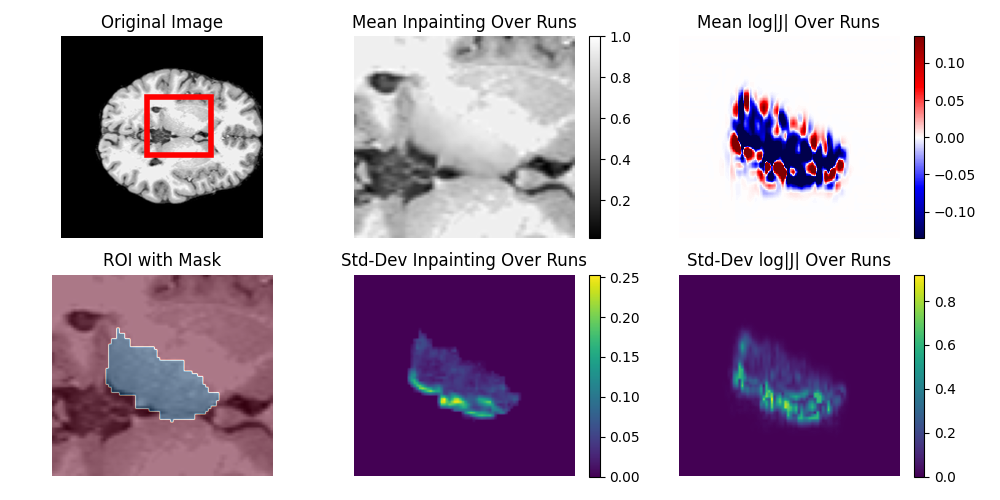}
    \caption{An example of the mean and standard deviation of the pixel intensities and $\logdetJ$ when inpainting a specific structure (Thalamus) over 10 runs on a specific subject.}
    \label{fig:example_measures}
\end{figure}

\section{Results}
\subsection{Accuracy and Consistency of Inpainting}

Our initial exploration characterises the consistency of inpainting over random seeds, i.e. drawing samples from the conditional distribution of $p(\image | \image \circ \mask)$. We illustrate the pixelwise distribution of our measures for one subject and a single structure in \figureref{fig:example_measures} and provide examples for additional structures in  \sectionref{sec:appendix_examples} of the appendix, where we include results for independently masking each hemisphere or bilateral masking. 

 \begin{figure}
     \centering
     \includegraphics[width=0.6\textwidth]{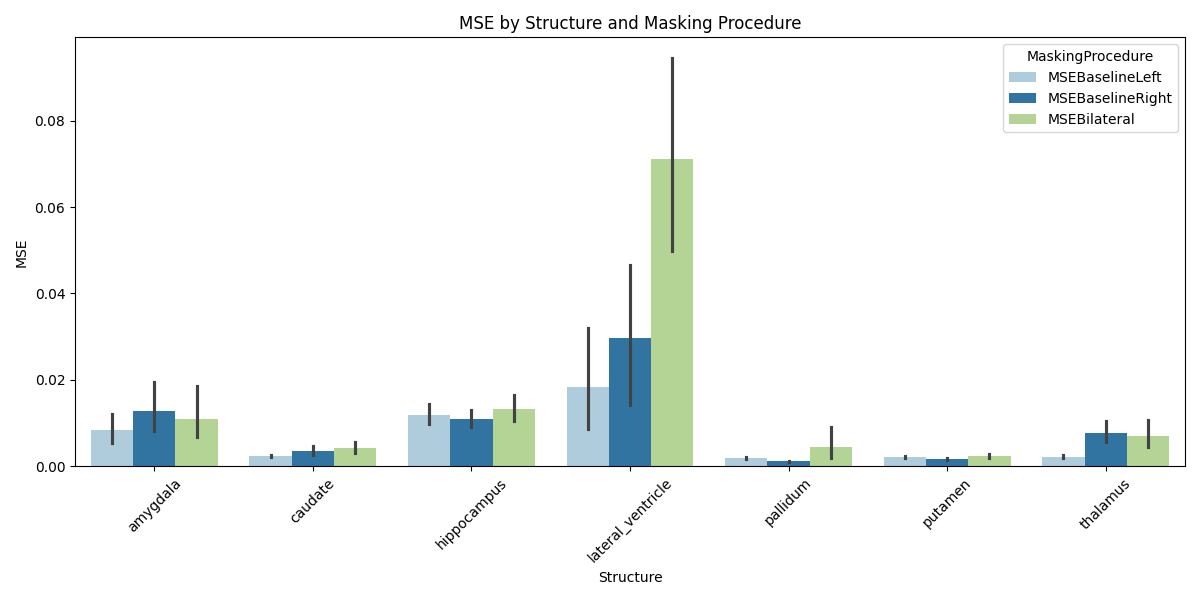}
     \includegraphics[width=0.6\textwidth]{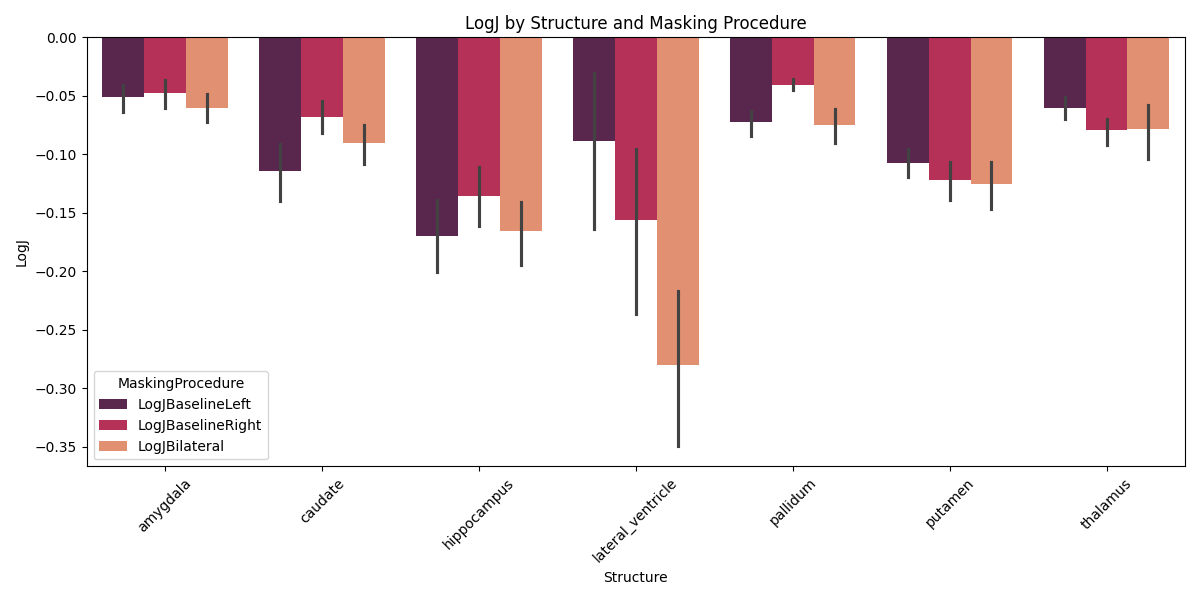}
     \caption{Plot of the mean and standard deviation of MSE (top) and $\logdetJ$ (bottom) across 20 inpainting runs on a single subject.}
     \label{fig:barplots}
\end{figure}

The average accuracy and between-run variability of inpainting is described in \figureref{fig:barplots} for a single subject and 20 inpainting runs. 

We  see the effect of lateral symmetry on the resulting inpainting by comparing bilateral masking with the baseline. This is also qualitatively illustrated in \sectionref{sec:appendix_examples} of the appendix where we include pixelwise maps of bilateral masking.

% \subsection{The effect of local context}
% Show violin plots of $\mathrm{mean}(\logdetJ)$ and $\imageDist$ over runs for a single subject with the baseline, bilateral and dilated masking procedures applied.

\subsection{Variance Analysis}
 \begin{figure}
     \centering
     \includegraphics[width=0.48\textwidth]{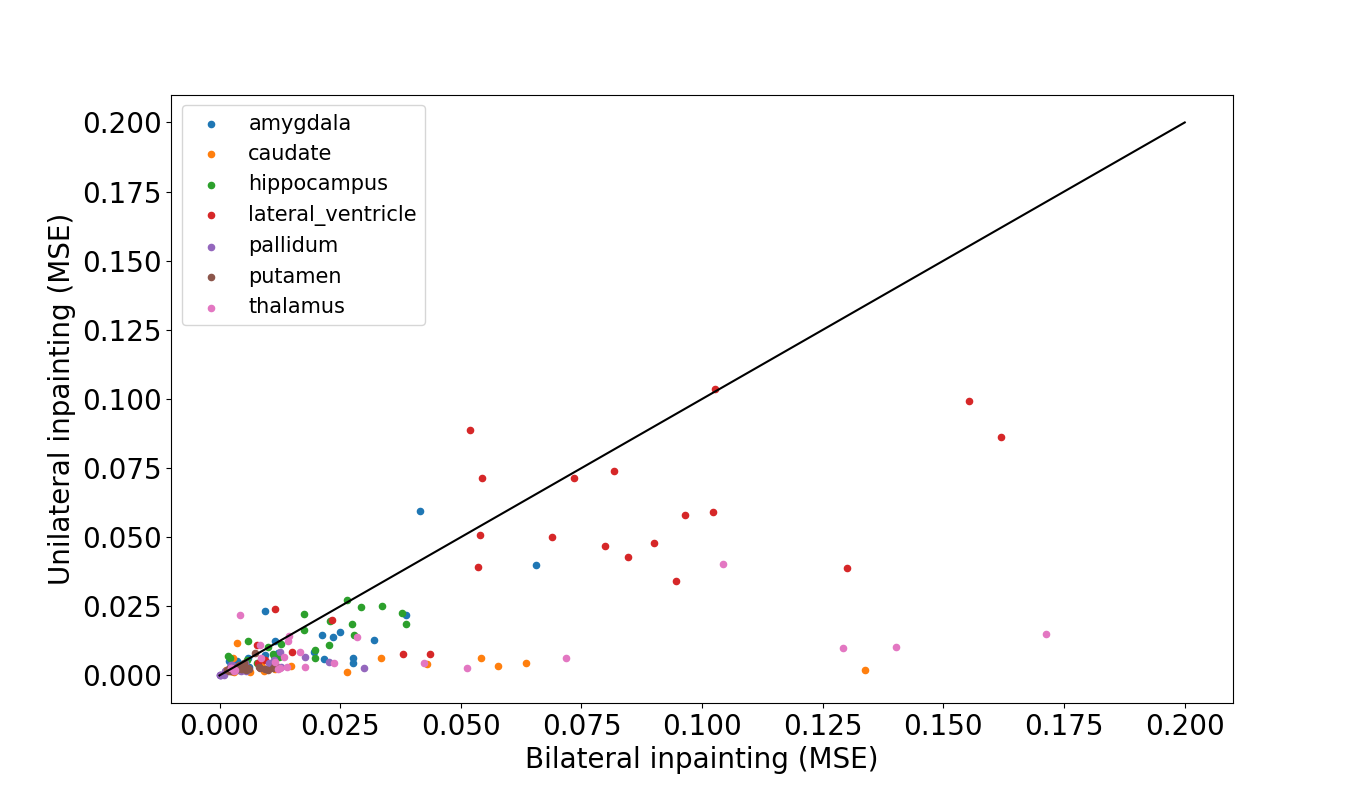}
     \includegraphics[width=0.48\textwidth]{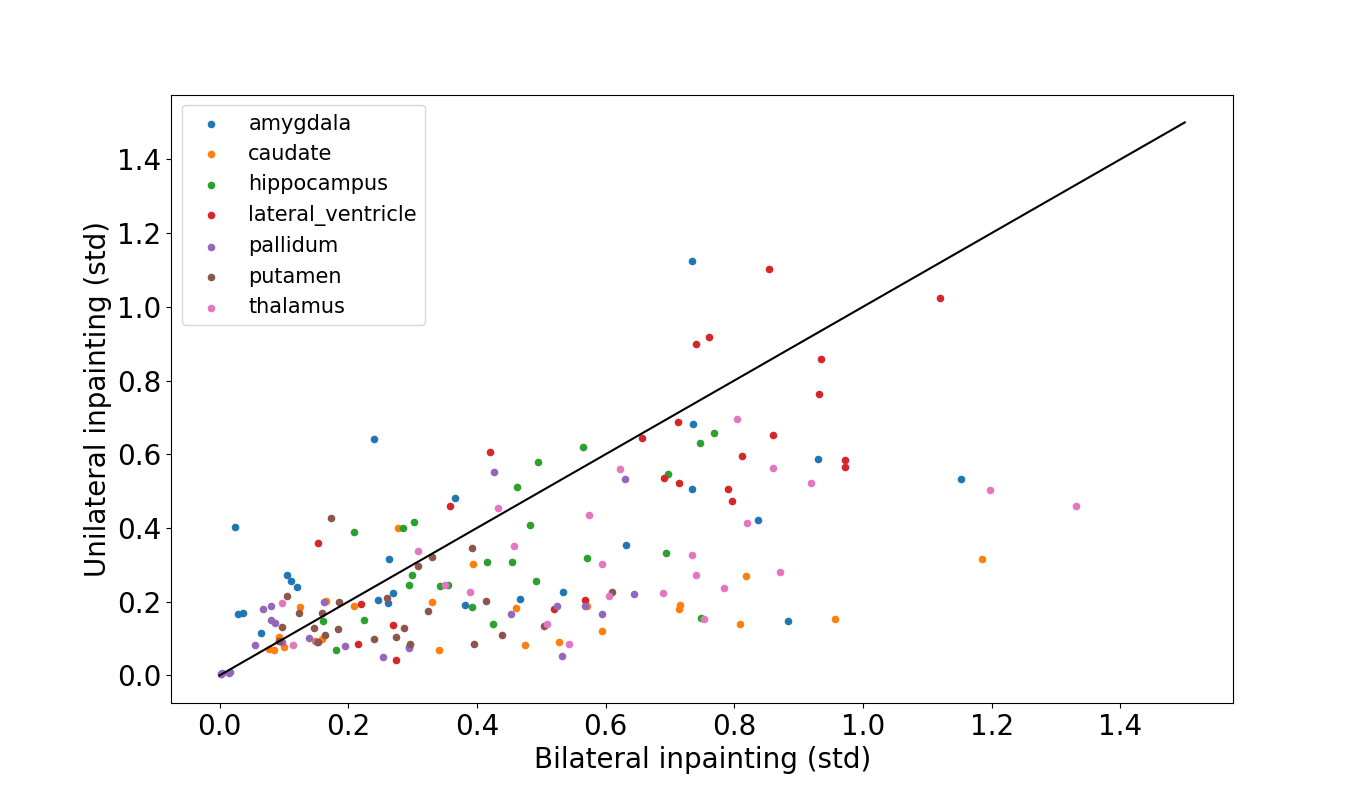}
     \caption{Left: a scatter plot of inpainting mean squared error (MSE) per-run per-subject for unilateral and bilateral masking. Right: a scatter plot of the standard deviation of the pixelwise $\logdetJ$ per-run per-subject.}
     \label{fig:scatter}
\end{figure}

We further investigate the introduced error and variability due to lateral symmetry across subjects. \figureref{fig:scatter} illustrates the increase in the level and variability of MSE and pixelwise variability in $\logdetJ$ with bilateral masking. We find that all structures seem to exhibit a strong dependence on lateral symmetry, and this seems to be particularly noticeable for  the lateral ventricle and Thalamus, which may be due to their relatively close proximity in the image.

\section{Discussion}
Our investigations have identified that the inpainting model that we tested demonstrated a systematic dependence on lateral symmetry. Further investigations will evaluate how this compares with a control of random masking, or more localised masking. 

A limitation of this work is we only evaluate a single, 2D axial-slice based, inpainting approach. As the brain is a 3D object, we would anticipate that conditioning on the 3D structure would be more informative than 2D. Future work will investigate this with 3D inpainting models, for example using \cite{graham2023unsupervised}.

This work has investigated training and evaluating a diffusion-based inpainting model on healthy brains. Future work could investigate the effects of training using a broader population in terms of age and disease status, where this distribution is characterised by the model. It was also be informative to evaluate the effects of these models on pathological data.

To measure the effects of inpainting we have chosen two measures, the mean squared error and the change of volume as characterised through non-linear registration. Future work could also incorporate structured similarity index measures (SSIM)~\cite{wang2004image} and could investigate the effect of hyperparameter choices, e.g. regularisation, for non-linear registration.

\section{Conclusion}
In this paper we have provided an initial investigation into the role of bilateral symmetry for inpainting brain MRI using the repaint algorithm in combination with a diffusion generative model. Our exploration indicates that masking the contra-lateral region has a substantial effect.

\bibliography{refs}
\newpage
\appendix

\section{Additional Examples\label{sec:appendix_examples}}
\begin{figure}[htbp]
\floatconts
  {fig:caudate}
  {An example of the mean and standard deviation of the pixel intensities and $\logdetJ$ when inpainting the Caudate over 10 runs on one subject. The top figure is the left side, middle the right side and the bottom is both sides.}
  {
     \centering
         \includegraphics[width=0.75\textwidth]{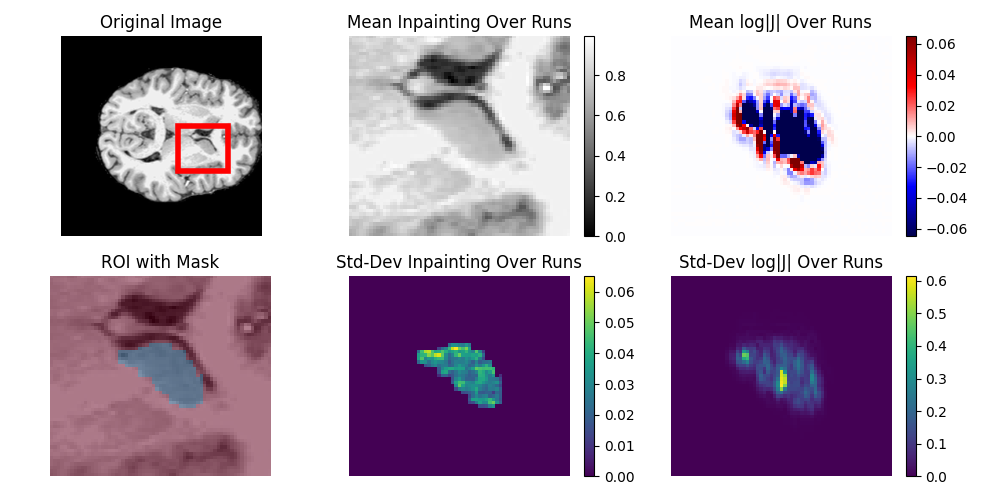}
         \includegraphics[width=0.75\textwidth]{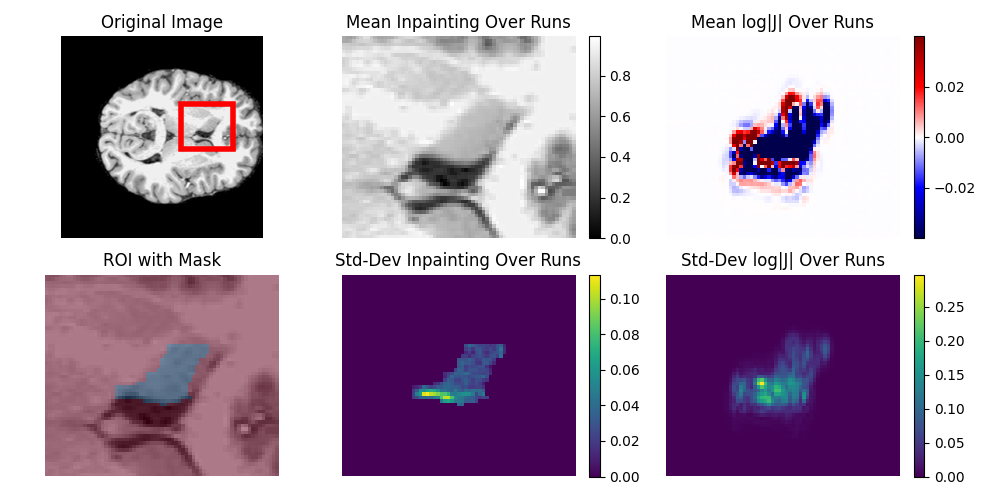}
         \includegraphics[width=0.75\textwidth]{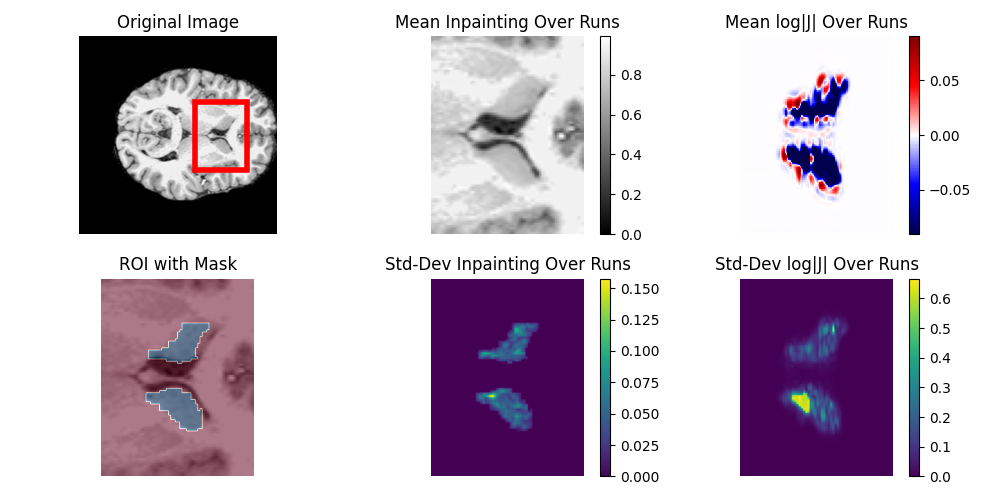}
}
\end{figure}

\begin{figure}[h!]
     \centering
         \includegraphics[width=0.8\textwidth]{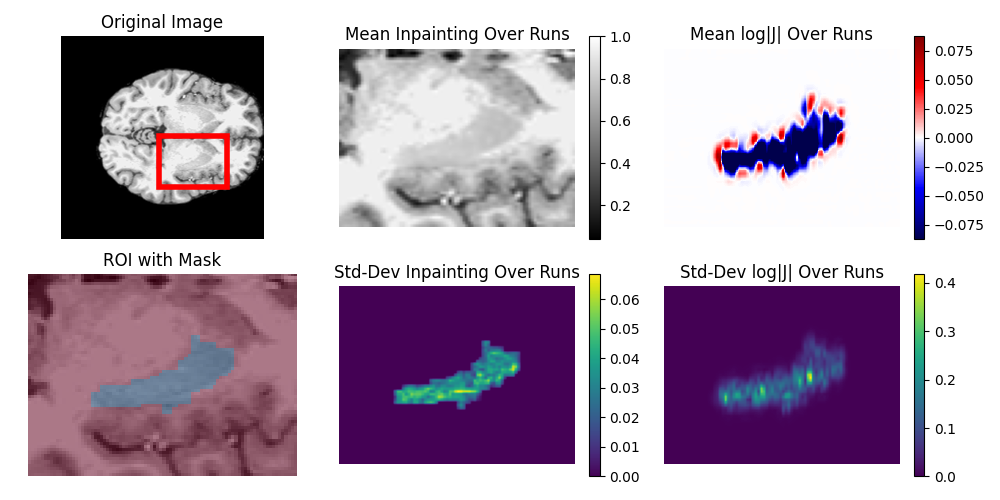}
         \includegraphics[width=0.8\textwidth]{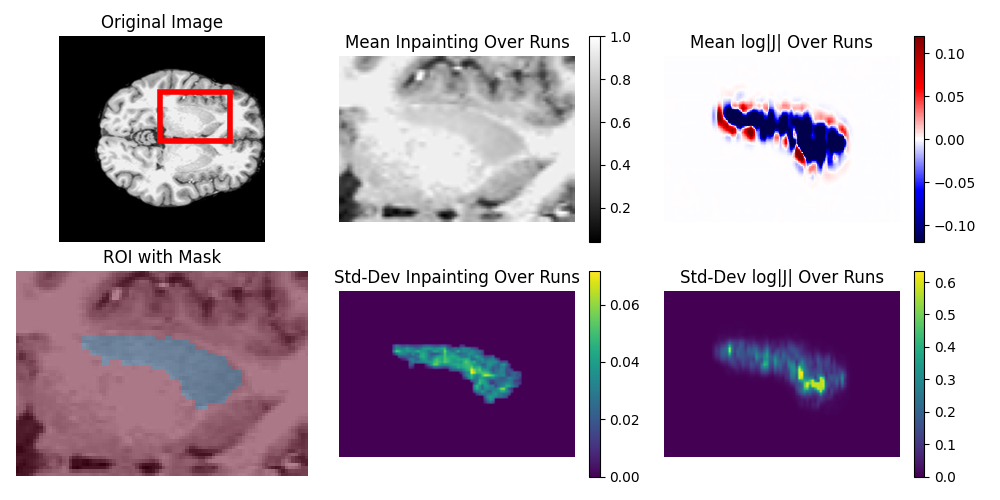}
         \includegraphics[width=0.8\textwidth]{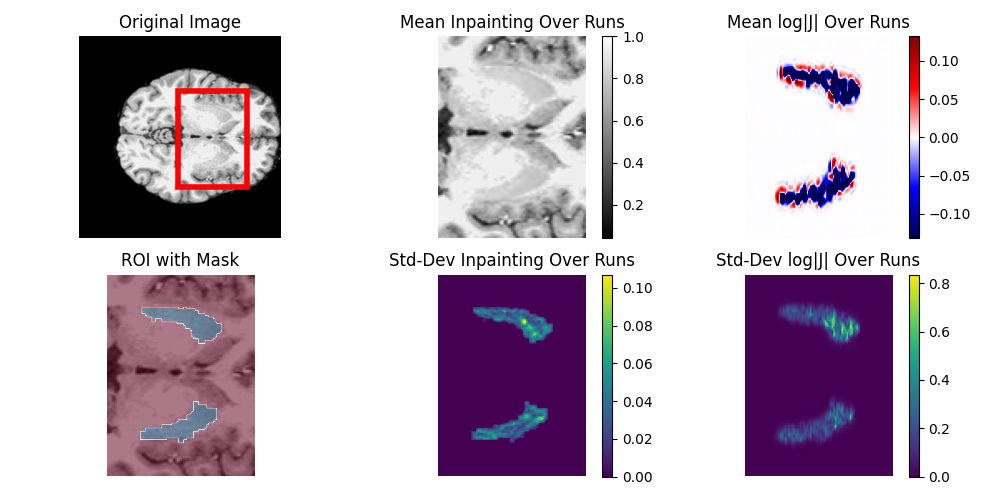}
    \caption{An example of the mean and standard deviation of the pixel intensities and $\logdetJ$ when inpainting the putamen over 10 runs on one subject. The top figure is the left side, middle the right side and the bottom is both sides.}
    \label{fig:caudate}
\end{figure}

\begin{figure}[h!]
     \centering
         \includegraphics[width=0.8\textwidth]{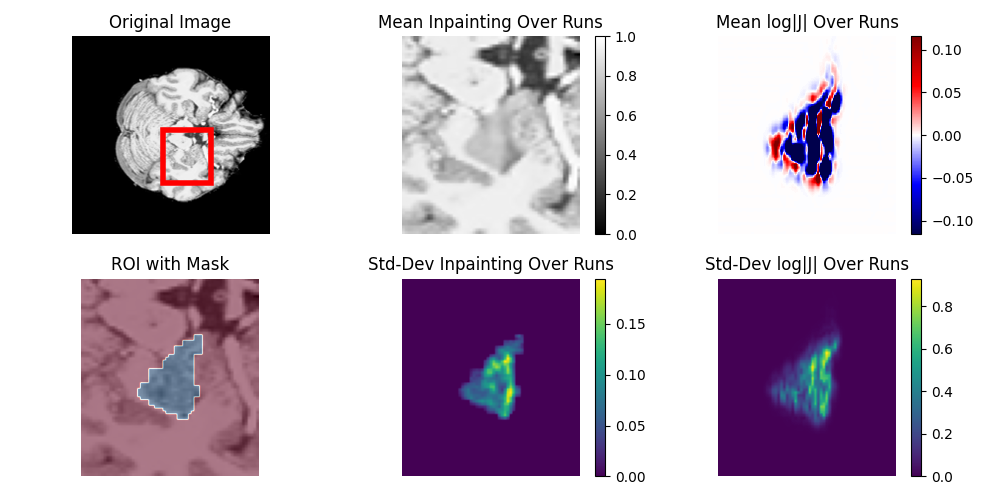}
         \includegraphics[width=0.8\textwidth]{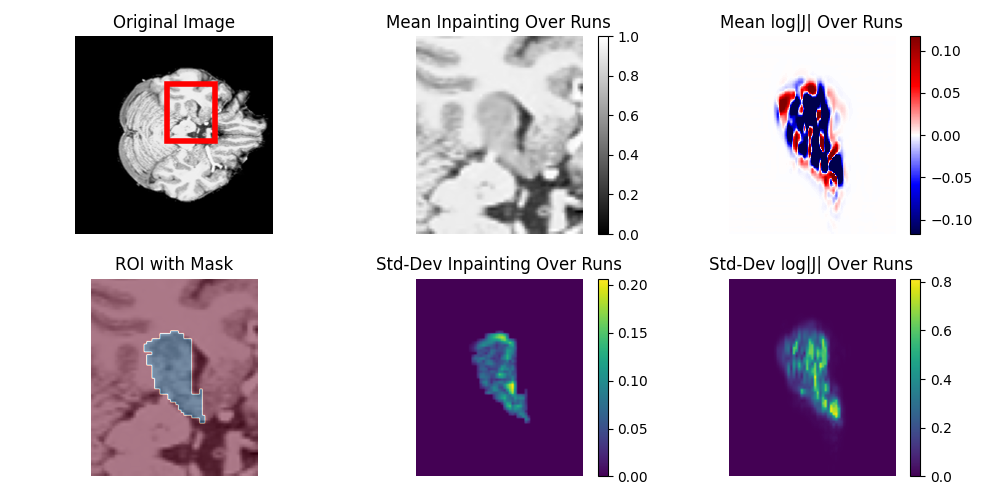}
         \includegraphics[width=0.8\textwidth]{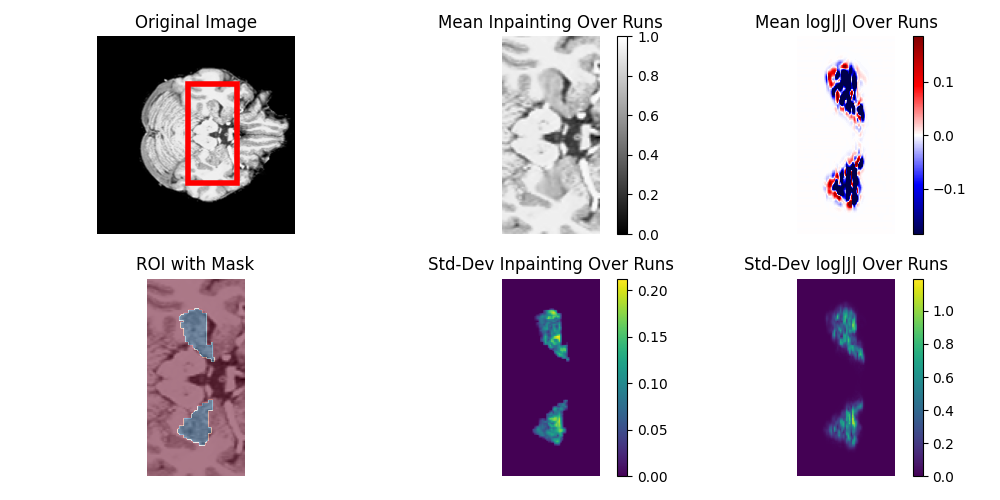}
    \caption{An example of the mean and standard deviation of the pixel intensities and $\logdetJ$ when inpainting the Hippocampus over 10 runs on one subject. The top figure is the left side, middle the right side and the bottom is both sides.}
    \label{fig:caudate}
\end{figure}

\begin{figure}[h!]
     \centering
         \includegraphics[width=0.8\textwidth]{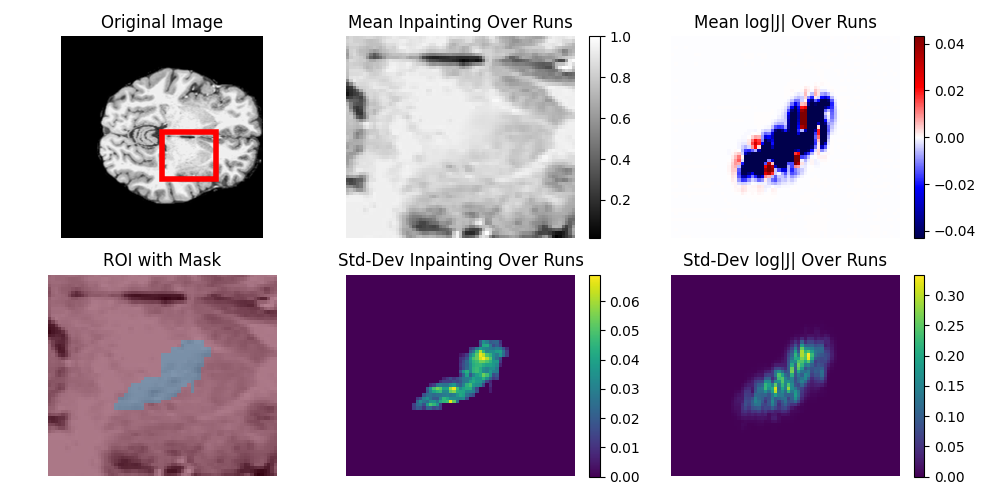}
         \includegraphics[width=0.8\textwidth]{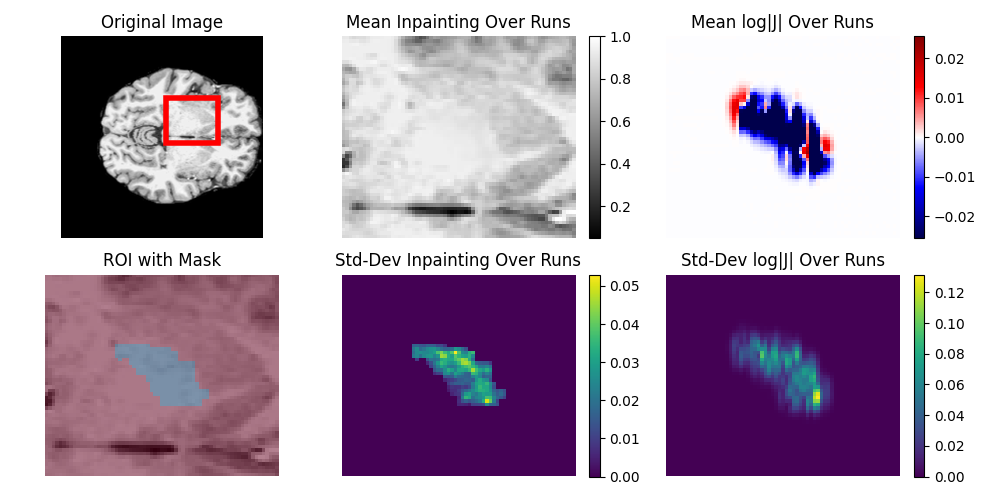}
         \includegraphics[width=0.8\textwidth]{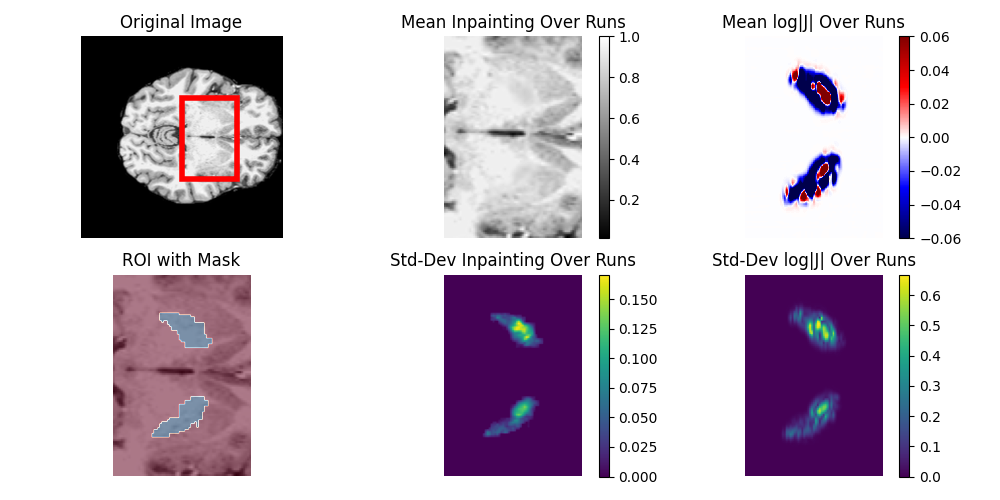}
    \caption{An example of the mean and standard deviation of the pixel intensities and $\logdetJ$ when inpainting the pallidum over 10 runs on one subject. The top figure is the left side, middle the right side and the bottom is both sides.}
    \label{fig:pallidum}
\end{figure}

\begin{figure}[h!]
     \centering
         \includegraphics[width=0.8\textwidth]{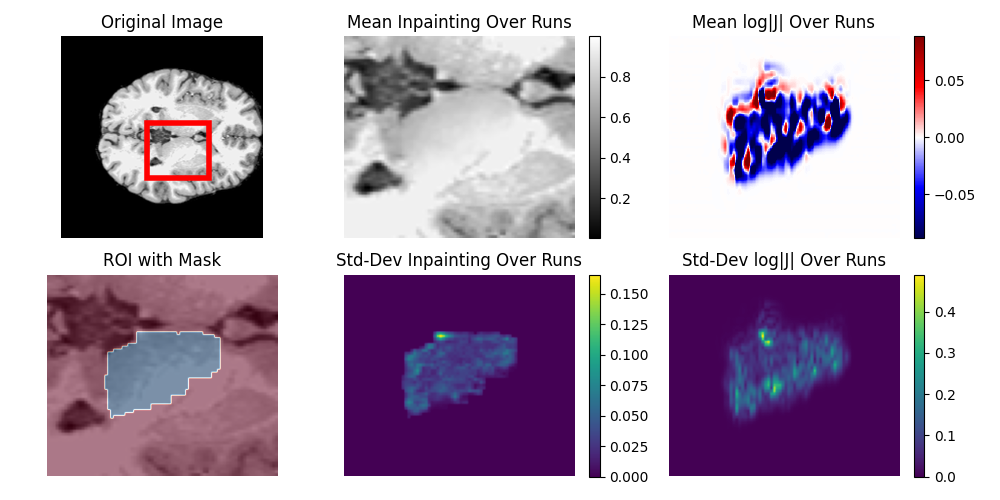}
         \includegraphics[width=0.8\textwidth]{figures/mean_std/right_thalamus.png}
         \includegraphics[width=0.8\textwidth]{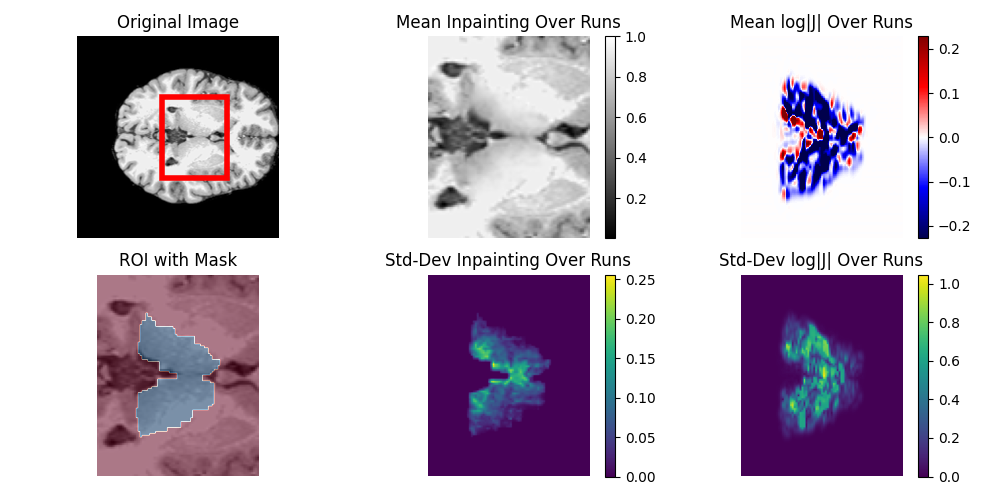}
    \caption{An example of the mean and standard deviation of the pixel intensities and $\logdetJ$ when inpainting the thalamus over 10 runs on one subject. The top figure is the left side, middle the right side and the bottom is both sides.}
    \label{fig:thalamus}
\end{figure}

\begin{figure}[h!]
     \centering
         \includegraphics[width=0.8\textwidth]{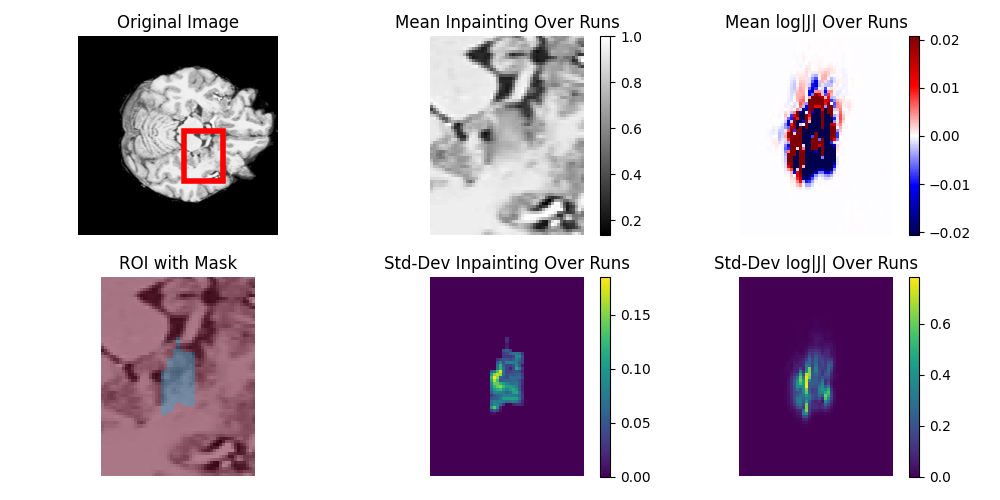}
         \includegraphics[width=0.8\textwidth]{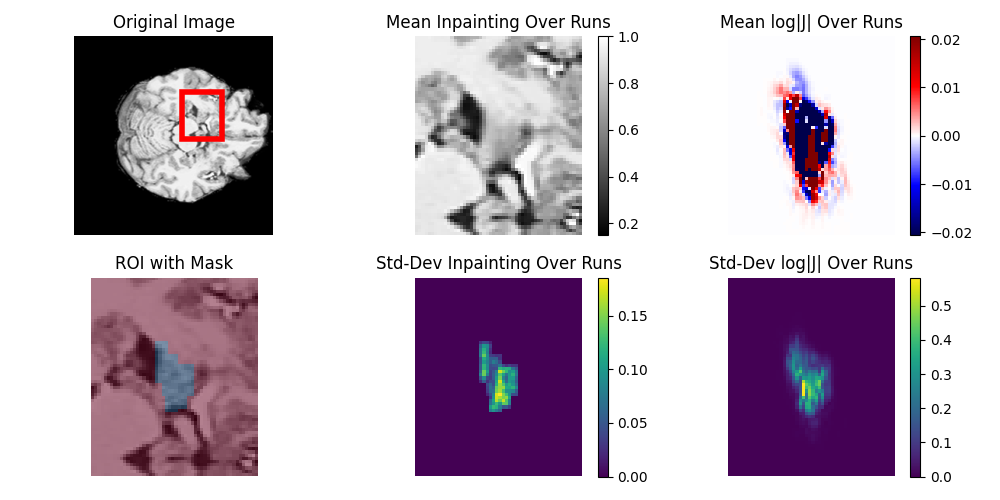}
         \includegraphics[width=0.8\textwidth]{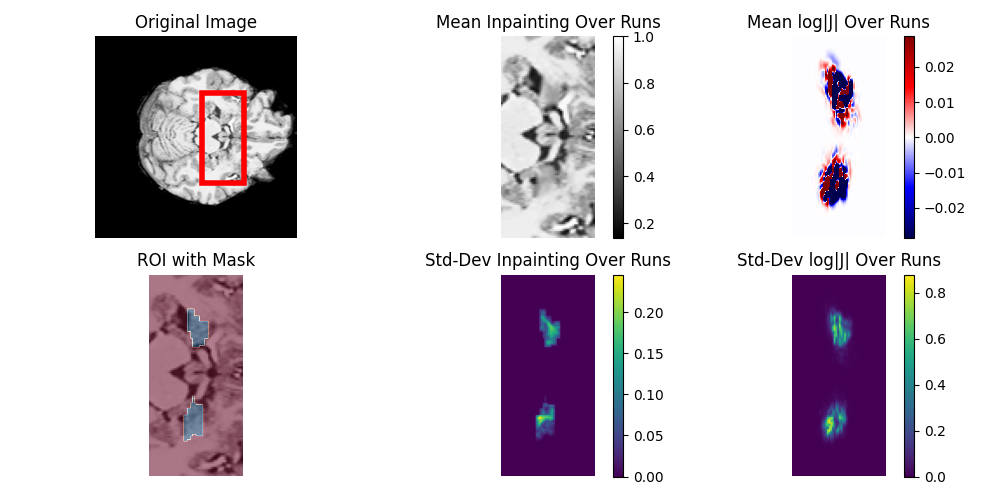}
    \caption{An example of the mean and standard deviation of the pixel intensities and $\logdetJ$ when inpainting the amygdala over 10 runs on one subject. The top figure is the left side, middle the right side and the bottom is both sides.}
    \label{fig:amygdala}
\end{figure}

\section{Registration Parameters}
We use Niftyreg with an in-plane grid spacing of $2.5 \times 2.5$ pixels. We remove all the bending energy and linear elasticity regularisation "-be 0.000 -le 0.000".
We use a cost-function of sum-of-squared differences.

"-sx 2.5 -sz 2.5 -sz 1  --ssd -ln 1 -nopy"

\end{document}